\documentclass[10pt,twocolumn,letterpaper]{article}

\usepackage[pagenumbers]{cvpr}
\usepackage{amsmath}
\usepackage{multirow}

\newcommand{\sd}[1]{{\tiny$\pm$#1}}

\definecolor{cvprblue}{rgb}{0.21,0.49,0.74}
\usepackage[breaklinks,colorlinks,allcolors=cvprblue]{hyperref}

\def\confName{CVPR}
\def\confYear{2027}

\title{Which Modality Decides?\\Counterfactual Modality Attribution for Multimodal Classifiers}

\author{
Vahidin Hasi\'{c}$^{1}$\quad
Chao Wang$^{2}$\quad
Luis C. Garcia-Peraza-Herrera$^{2}$\quad
David Watson$^{2,\dagger}$\quad
Senka Krivic$^{1,\dagger}$\\[-2pt]
{\small $^{1}$Faculty of Electrical Engineering, University of Sarajevo, Bosnia and Herzegovina\quad
$^{2}$King's College London, United Kingdom}\\[-4pt]
{\footnotesize\tt vahidin.hasic@etf.unsa.ba, chao.wang@kcl.ac.uk, luis\_c.garcia\_peraza\_herrera@kcl.ac.uk,}\\[-5pt]
{\footnotesize{\tt david.watson@kcl.ac.uk, senka.krivic@etf.unsa.ba}\quad$^{\dagger}$Equal contribution}
}

\makeatletter
\patchcmd{\@maketitle}{\vspace*{12pt}}{}{}{\errmessage{author spacing patch failed}}
\patchcmd{\@maketitle}{\vspace*{24pt}}{\vspace*{20pt}}{}{\errmessage{title spacing patch failed}}
\patchcmd{\@maketitle}{\vskip .375in}{\vskip .25in}{}{\errmessage{top spacing patch failed}}
\makeatother

\begin{document}
\maketitle

\begin{abstract}
Multimodal classifiers can achieve high accuracy while exploiting modality-specific shortcuts. Modality attribution quantifies the contributions of image and text, but does not by itself identify which input changes would alter a suspicious prediction. We introduce \textbf{Counterfactual Modality Attribution (CMA)}, which connects modality-level contributions to explicit prediction-changing edits. CMA uses image and text diffusion priors to generate counterfactual references and constructs a two-player reference game whose Shapley–Taylor decomposition separates image, text, and interaction contributions. The method requires class-score queries but no model gradients. Attribution profiles are compared with a reference population to prioritize predictions with atypical modality reliance for review, while counterfactual edits expose localized evidence for inspection. On controlled image- and text-biased classifiers, CMA identifies the ground truth dominant modality for all evaluated inputs. On planted-shortcut benchmarks, it achieves shortcut-reliance detection AUROCs of 0.987 on Synthetic MNIST and 1.000 on OpenI, with image-shortcut localization IoUs of 0.97 and 0.84, respectively. An API experiment demonstrates applicability to a native multimodal language model with scored answer choices. CMA enables predictions to be screened by modality reliance and investigated through explicit counterfactual edits.

\end{abstract}

\section{Introduction}
\label{sec:introduction}

\begin{figure}[!t]
    \centering
    \includegraphics[width=\linewidth]{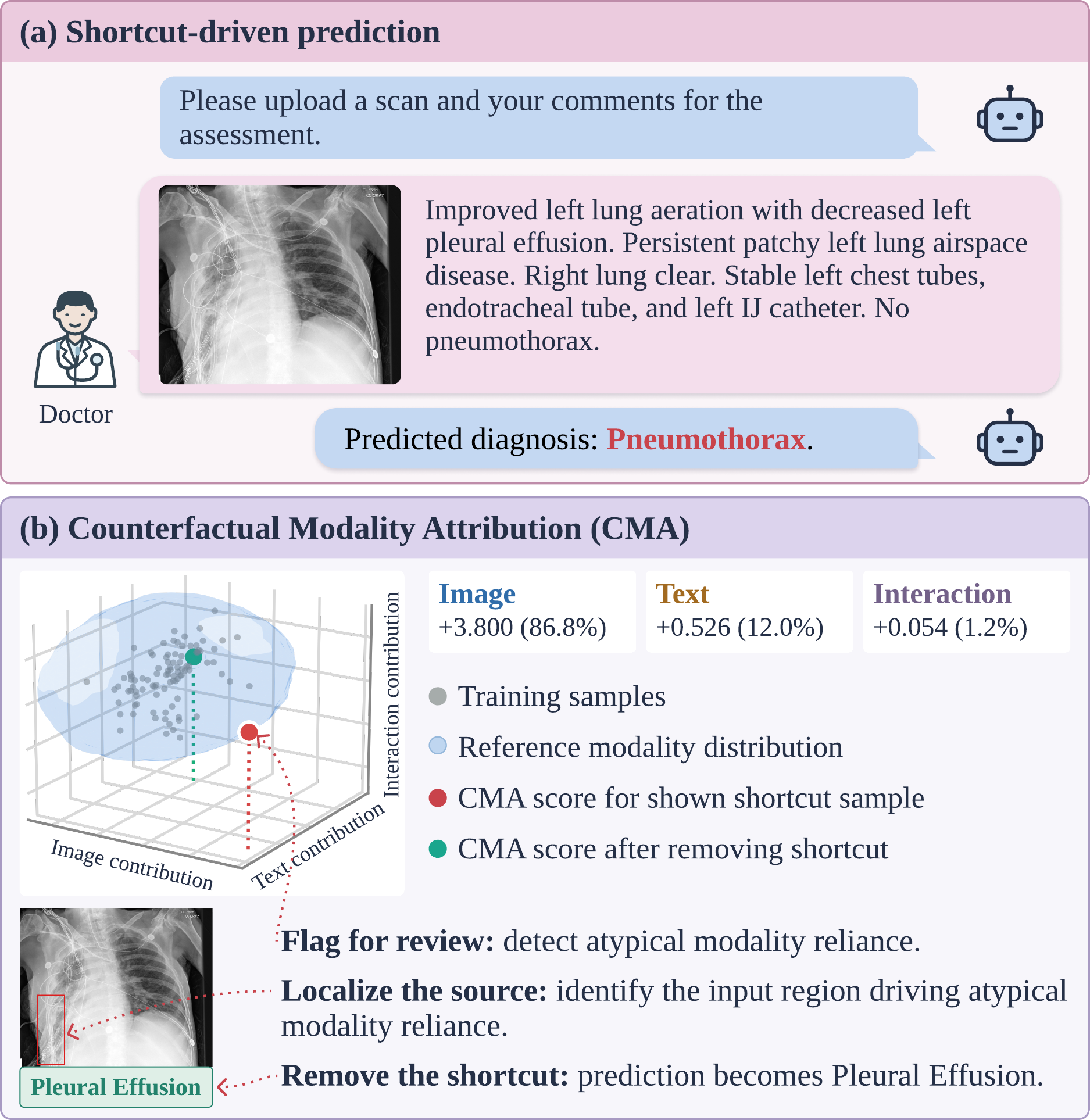}
    \caption{
   \textbf{Illustration of modality attribution failure.} \textbf{(a)} A multiclass prediction of \emph{pneumothorax} disagrees with report negation.
\textbf{(b)} CMA attributes $86.8\%$, $12\%$, and $1.2\%$ of attribution to image, text, and interaction, respectively.
The unnormalized Shapley--Taylor attribution is atypical relative to the training reference population (red), flagging the prediction for review. Removing the highlighted tube-related shortcut changes the prediction to \emph{pleural effusion} and yields a typical profile (green), illustrating how atypical modality reliance can guide inspection of a localized shortcut.
    }
    \label{fig:motivation_image}
\end{figure}

Multimodal large language models (MLLMs) combine complementary evidence from multiple modalities. In clinical decision support, for example, radiologists interpret radiological images alongside patient histories and physician notes~\citep{Acosta2022,Soenksen2022}. However, access to both modalities does not ensure that a model uses relevant evidence for a given prediction. A classifier may exploit a visual shortcut of a diagnosis or follow misleading text despite contradictory image evidence~\citep{buckley2026multimodal}. Such dependencies can remain hidden in model evaluation, raising an important unanswered question: \emph{which modality drives a prediction, and what evidence within it drives the decision?}

Fig.~\ref{fig:motivation_image} illustrates a classifier that predicts pneumothorax from a tube-related image shortcut despite a report negation.
Identifying the image as dominant narrows the investigation. However, it does not demonstrate an inappropriate reliance, since visual evidence may be decisive.
Assessing the prediction requires comparing attribution scores with the model's usual reliance patterns and intervening on the suspected shortcut. An atypical attribution profile flags the input for review, while a counterfactual provides concrete evidence for inspection. Together, they link modality-level attribution to input features that influence the decision.

Existing explanation methods address parts of this problem. Feature attributions identify influential input regions~\citep{selvaraju2017grad,benmelech2024lvlm,li2025tam}. MM-SHAP measures relative modality contributions~\citep{parcalabescu2023mm}, and InterSHAP separates modality effects from cross-modal interactions~\citep{wenderoth2025measuring}. These contributions depend on the contrasts used to evaluate the model. Masking features or replacing a modality measures the model's response to absent or unrelated evidence, but a modality score alone does not identify the content responsible for a suspicious decision. The missing connection is between a modality-level attribution and a prediction-changing counterfactual that can be inspected in the input.

We introduce \textbf{Counterfactual Modality Attribution (CMA)}, a framework for explaining multimodal image--text LLM classifiers through generated counterfactuals. CMA uses image and text diffusion priors to search for nearby counterfactuals under joint editing constraints. The resulting references define a two-player cooperative game over the modalities. A second-order Shapley--Taylor decomposition separates image, text, and interaction contributions to the model output. CMA requires only class-score queries, without classifier gradients or model weights, enabling its application to native MLLMs with scored answer choices.

Each input's attribution is compared with a reference population and unusual profiles are ranked for review. The counterfactual edits then indicate which image regions or text passages can change the flagged prediction. This procedure is useful when a shortcut produces a departure from the model's usual modality reliance. It does not assume that balanced attribution is desirable, or that every unusual profile is erroneous. Instead, it provides a quantitative basis for selecting cases and concrete evidence for examining them.

We evaluate CMA on controlled multimodal classifiers and planted shortcuts with known locations in Synthetic MNIST and OpenI. These experiments assess modality identification, counterfactual quality, shortcut detection, and localization. Progressive corruption examines attribution under degraded inputs, and an API experiment demonstrates applicability to a native MLLM.

Our contributions are:
\begin{itemize}
\item  \textbf{Counterfactual modality attribution.} We define a counterfactual reference game that decomposes prediction into image, text, and interaction contributions.
\item \textbf{Attribution-guided shortcut inspection.} We connect attribution scores to counterfactual edits to prioritize predictions for review and localize candidate shortcuts within the responsible modality.
\item \textbf{Evaluation against measured model behavior.}
We validate attributions against independent modality interventions and known shortcut locations.
\end{itemize}

\section{Related Work}
\label{sec:related_work}

Existing work on explainability for multimodal models spans feature attribution, counterfactual explanations, and diffusion-based models.%

\textbf{Modality contributions and interactions.}
Existing methods quantify modality reliance at the performance and prediction levels. Perceptual Score~\citep{gat2021perceptual} measures performance changes under modality permutation, SHAPE~\citep{hu2022shape} uses performance-based Shapley scores to assess modality contributions and cooperation, and EMAP~\citep{hessel2020does} tests the predictive value of cross-modal interactions through an additive projection. At the prediction level, MM-SHAP~\citep{parcalabescu2023mm} aggregates token- and patch-level Shapley values into modality contributions, while InterSHAP~\citep{wenderoth2025measuring} decomposes modality contributions and interactions locally and globally. MultiSHAP~\citep{wang2026multishap} explains fine-grained patch--token interactions through black-box evaluations. CMA instead defines a modality game relative to generated joint counterfactuals: its Shapley--Taylor decomposition separates prediction changes into image, text, and interaction terms, while the edits provide inspectable evidence of what changes the decision.

\textbf{Feature attribution and multimodal interpretation.}
LVLM-Interpret~\citep{benmelech2024lvlm} visualizes attention and relevance between generated text and image patches, while TAM~\citep{li2025tam} produces token-specific visual explanations by reducing interference from preceding context. EAGLE~\citep{chen2026mllms} uses black-box sufficiency and necessity objectives to identify regions supporting output tokens and assess reliance on visual evidence versus language priors. MultiViz~\citep{liang2022multiviz} examines unimodal importance, cross-modal interactions, representations, and predictions. These methods connect outputs to features or representations, helping reveal failures of visual grounding.
CMA complements them by linking a modality-level attribution profile to explicit image--text counterfactuals where unusual modality attribution signals suspicious predictions that are prioritized for review.

\textbf{Counterfactual explanations and generative priors.}
Counterfactual explanations seek small input changes that alter a model's prediction~\citep{wachter2018counterfactual,guidotti2024}. DiME~\citep{jeanneret2022diffusion}, DVCE~\citep{augustin2022diffusion}, and ACE~\citep{jeanneret2023adversarial} generate visual counterfactuals using diffusion models. FastDiME~\citep{weng2024fast} accelerates generation and uses edits to investigate image shortcuts, while BTTF~\citep{wang2026back} extends generation to video. For text, MiCE~\citep{ross2021mice} searches for minimal prediction-changing edits, and Polyjuice~\citep{wu2021polyjuice} generates controlled perturbations for model analysis. Feynman--Kac steering~\citep{singhal2025general} provides a complementary generation mechanism, using intermediate potentials and particle resampling to guide diffusion at inference time. CMA builds on this principle with DDPM image priors~\citep{ho2020ddpm} and an absorbing-state masked text diffusion prior following~\citet{sahoo2024simple}.

\begin{figure*}[t!]
    \centering
    \includegraphics[width=\linewidth]{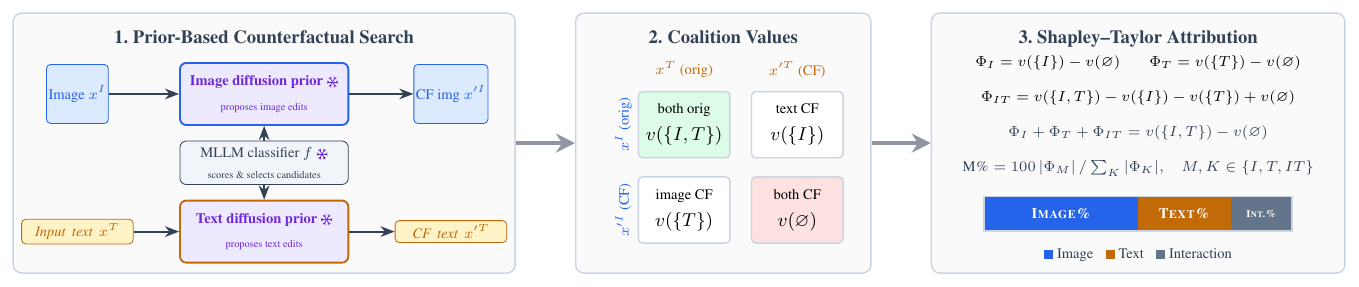}
    \caption{
   \textbf{Overview of CMA.}
\textbf{(1)} Independently pretrained image and text diffusion priors propose edits, scored and selected using the MLLM classifier without access to its gradients.
\textbf{(2)} Original and counterfactual modalities form four coalitions in a two-player game.
\textbf{(3)} Shapley--Taylor decomposition quantifies player modality attribution into separate image, text, and interaction contributions.
Attribution scores reveal which modality drives the model's prediction with evidence, which when compared with a training reference population flags unusual cases for human review.
    }
    \label{fig:proposed_method}
\end{figure*}

\section{Background}
\label{sec:background}

\textbf{Image diffusion.}
A denoising diffusion probabilistic model (DDPM)~\citep{ho2020ddpm} defines a Gaussian corruption process for a clean image $x_0^I$:
$
q(x_s^I\mid x_0^I)=
\mathcal{N}\!\left(
\sqrt{\bar\alpha_s}\,x_0^I,
(1-\bar\alpha_s)\mathbf{I}
\right),
$
where $\alpha_s=1-\beta_s$, $\bar\alpha_s=\prod_{r=1}^{s}\alpha_r$, and $\{\beta_s\}_{s=1}^{T}$ is the noise schedule. The noise predictor $\epsilon_\theta(x_s^I,s)$ gives the clean-image estimate
\begin{equation*}
\hat x_0^I=
\frac{x_s^I-\sqrt{1-\bar\alpha_s}\,\epsilon_\theta(x_s^I,s)}
{\sqrt{\bar\alpha_s}}
\end{equation*}
and parameterizes the Gaussian reverse-transition mean as
\begin{equation*}
\mu_\theta(x_s^I,s)=\frac{1}{\sqrt{\alpha_s}}
\left(x_s^I-
\frac{\beta_s}{\sqrt{1-\bar\alpha_s}}\,
\epsilon_\theta(x_s^I,s)\right).
\end{equation*}
Denoising a partially corrupted input enables image editing~\citep{meng2022sdedit}, while reinserting appropriately noised observations outside an editing mask constrains the retained region~\citep{lugmayr2022repaint}. An additional noise--denoise pass provides a stochastic reconstruction, as in diffusion purification~\citep{nie2022diffusion}.

\textbf{Text diffusion.}
Absorbing-state masked diffusion~\citep{sahoo2024simple} progressively replaces sequence elements with \texttt{[MASK]}. We represent record lines or report sentences as categorical units in fixed slots, using a vocabulary $\mathcal{U}$ of training-corpus units and \texttt{[EMPTY]}, an unoccupied-slot marker distinct from \texttt{[MASK]}. At time $t\in[0,1]$, each original unit is retained with probability $\alpha(t)$ under a decreasing schedule. The denoiser $p_\theta(u_l\mid\tilde{x}_t^T,t)$ predicts the clean unit at slot $l$. Reverse sampling progressively unmasks slots, preserving revealed units and keeping unselected slots fixed during constrained editing.

\textbf{Sampling.}
Sequential Monte Carlo (SMC) approximates a sequence of distributions using weighted particles and resampling~\citep{wu2023practical}. For a denoising trajectory $z_{s:T}$, consider unnormalized targets $\gamma_s(z_{s:T})=p_{\mathrm{prior}}(z_{s:T})G_s(z_s)$, where $G_s>0$ is a steering potential. Propagating a particle from $s+1$ to $s$ with proposal $r_s(z_s\mid z_{s+1})$ gives the incremental importance factor
\begin{equation}
\omega_s=
\frac{p_{\mathrm{prior}}(z_s\mid z_{s+1})}
{r_s(z_s\mid z_{s+1})}
\frac{G_s(z_s)}{G_{s+1}(z_{s+1})}.
\label{eq:smc_increment}
\end{equation}
When the proposal matches the prior transition, only the potential ratio remains. Resampling by normalized weights concentrates computation on trajectories favored by  target.

\paragraph{Shapley--Taylor.}
Let $v:2^N\to\mathbb{R}$ be a cooperative game with modalities $N=\{I,T\}$ as players. For two players, the order-two Shapley--Taylor index~\citep{dhamdhere2020shapley} equals the nonempty M\"obius coefficients, or Harsanyi dividends:
\begin{equation*}
\begin{aligned}
\Phi_I = v(\{I\})-v(\emptyset), \quad
\Phi_T = v(\{T\})-v(\emptyset),\\
\Phi_{IT} = v(\{I,T\})-v(\{I\})-v(\{T\})+v(\emptyset).
\end{aligned}
\end{equation*}
The main effects measure each modality's contribution when added alone to the reference coalition, while the interaction measures the residual departure from additivity. The decomposition is exact:
\begin{equation}
\Phi_I+\Phi_T+\Phi_{IT}
=v(\{I,T\})-v(\emptyset).
\label{eq:stii_efficiency}
\end{equation}
Conventional Shapley values allocate half the interaction to each modality, giving $\theta_I=\Phi_I+\tfrac12\Phi_{IT}$ and $\theta_T=\Phi_T+\tfrac12\Phi_{IT}$. CMA retains the three signed terms separately. Their values depend on the coalition score and reference construction. They are not probabilities and need not sum to one.

\textbf{Typicality ranks.}
A lower-tail rank compares a test input's typicality with reference scores, with smaller values indicating greater atypicality. Split conformal calibration~\citep{vovk2005algorithmic,bates2023testing} gives such ranks a marginal false-alarm guarantee when the scoring rule is fitted independently of held-out calibration inputs and calibration and test inputs are exchangeable conditional on that rule. Reusing the fitting population with leave-one-out distances does not satisfy these split-conformal conditions. CMA uses the latter protocol for empirical review ranking, as defined in Sec.~\ref{sec:method_tail}.

\section{Methodology}
\label{sec:methodology}

Let $f$ be a pretrained image--text classifier with class probabilities $p_f(c\mid x)$ for $x=(x^I,x^T)$ and $c\in\{1,\ldots,C\}$. The image is $x^I\in\mathbb{R}^{3\times H\times W}$, and $x^T$ is the tokenized text. The prediction $\hat y=\operatorname*{arg\,max}_c p_f(c\mid x)$ is explained by keeping $\hat y$ fixed throughout generation and attribution. CMA constructs counterfactuals, decomposes the resulting score contrast into modality and interaction terms, and uses attribution profiles to prioritize predictions for review (Fig.~\ref{fig:proposed_method}).

We use the one-versus-rest log-odds of $\hat y$:
\[
\ell_{\hat y}(\tilde x)
=\log\frac{p_f(\hat y\mid\tilde x)}{1-p_f(\hat y\mid\tilde x)}
=f_{\hat y}(\tilde x)-\log\!\sum_{c\ne\hat y}e^{f_c(\tilde x)}.
\]
This score compares $\hat y$ with all competing classes and is invariant to a common logit shift.

\subsection{Counterfactual Generation}
\label{sec:method_cf}

A joint counterfactual $x'=(x'^I,x'^T)$ changes the predicted class, $\operatorname*{arg\,max}_c p_f(c\mid x')\ne\hat y$, under a search that permits changes to both modalities. Joint search does not require both modalities to change in every accepted candidate. Independently trained image and text priors propose edits, while classifier scores steer candidate selection.

\textbf{Image proposals.}
Global editing starts from a partially noised input, $x^I_{s_0}\sim q(x^I_{s_0}\mid x^I)$, and follows the reverse kernel. For local editing, a binary mask $m$ specifies editable pixels. At step $s$, a prior proposal $u_{s-1}\sim p_\theta(\cdot\mid x_s^I)$ is combined with a noised observation $v_{s-1}\sim q(x_{s-1}^I\mid x^I)$:
\begin{equation*}
x_{s-1}^I=m\odot u_{s-1}+(1-m)\odot v_{s-1}.
\label{eq:local_edit}
\end{equation*}
The retained region follows the input's noise level, recovering the original image at step zero. Global editing uses $m=\mathbf{1}$; local masks constrain search without known shortcut locations. Spatially proposed windows are scored by the classifier.

\textbf{Text proposals.}
Text is edited as record lines or report sentences under the masked diffusion prior. Eligible slots are determined by the text structure, and classifier scores guide slot selection without requiring shortcut annotations. For a current candidate $z=(z^I,z^T)$ with a clean image $z^I$, let $\pi_l(u)=p_\theta(u\mid\tilde z_t^T,t)$ be the prior probability of unit $u$ at editable slot $l$, and let $\pi_l^\star=\max_{u\in\mathcal U}\pi_l(u)$. The prior-supported set
$\mathcal C_l=\{u\in\mathcal U:\pi_l(u)\geq\eta\pi_l^\star\}$ is retained with $\eta\in(0,1]$, and replacements are sampled from
\begin{equation*}
q_l(u\mid z)\propto
\left(\frac{\pi_l(u)}{\pi_l^\star}\right)^a
\left[1-p_f(\hat y\mid z^I,z^T_{l\leftarrow u})\right]^g,
\label{eq:textprop}
\end{equation*}
for $u\in\mathcal C_l$, with zero probability otherwise. The parameters $a,g\geq0$ control prior preference and classifier steering. Sequential replacements are scored in the updated context, while unselected slots remain unchanged.

\textbf{Particle steering.}
Joint search maintains particles comprising an image diffusion state and a text candidate. Text is proposed using the original image for classifier scoring, then fixed throughout each image trajectory; text masking time $t$ is therefore separate from image reverse steps $s$. At scoring step $s$, the clean-image estimate $\hat x_0^I(s)$ and candidate text $x'^T$ define
\begin{equation}
G_s=\left[1-p_f\!\left(\hat y\mid\hat x_0^I(s),x'^T\right)\right]^{\lambda_s},
\label{eq:twist}
\end{equation}
where $\lambda_s$ controls steering strength. Between scoring steps $s'$ and $s$, weights are multiplied by $G_s/G_{s'}$ and normalized for resampling, favoring trajectories that reduce the probability of $\hat y$~\citep{wu2023practical}. For $g>0$, text receives classifier guidance during both proposal and selection, while images receive it only through selection. This asymmetry can affect the joint references found. Exact importance sampling under the original priors would also require proposal-density corrections, including for text (Eq.~\ref{eq:smc_increment}). CMA instead uses this SMC-inspired update for heuristic search, without claiming exact conditional sampling.

\textbf{Acceptance and reference selection.}
We test whether prediction changes persist under stochastic image reconstruction:
$
P_\tau(z)=\mathcal D_{\theta,\tau}\!\left(
\sqrt{\bar\alpha_\tau}\,z+
\sqrt{1-\bar\alpha_\tau}\,\epsilon\right),
\label{eq:projection}
$
where $\epsilon\sim\mathcal N(0,\mathbf I)$ and $\mathcal D_{\theta,\tau}$ denoises from step $\tau$ to zero. A candidate with prediction $y'=\arg\max_c p_f(c\mid x')$ is accepted if $y'\ne\hat y$ and
\begin{equation*}
\arg\max_c p_f\!\left(c\mid P_\tau^{(b)}(x'^I),x'^T\right)=y',
\quad b=1,\ldots,B,
\end{equation*}
for $B$ independent reconstructions. This finite stability check does not certify semantic validity.
From $M>0$ distinct accepted candidates, we retain $R=\min(K,M)$ references without duplication, favoring distinct edited texts and ranking by $d_{\mathrm{prox}}(x,x')$, which combines normalized image change and an edited-text penalty. If none passes, we retain the candidate with the lowest $\ell_{\hat y}$ and flag the search as unsuccessful. This fallback supports score-contrast decomposition but is not an accepted counterfactual.

\subsection{Modality Attribution}
\label{sec:method_attr}
For each reference $x'_r=(x'^I_r,x'^T_r)$, coalition $S\subseteq\{I,T\}$ restores the original modalities in $S$:
\[
\begin{aligned}
v_r(\{I,T\})&=\ell_{\hat y}(x^I,x^T), &
v_r(\{I\})&=\ell_{\hat y}(x^I,x'^T_r),\\
v_r(\{T\})&=\ell_{\hat y}(x'^I_r,x^T), &
v_r(\emptyset)&=\ell_{\hat y}(x'^I_r,x'^T_r).
\end{aligned}
\]
All coalitions share the same reference. Mixed inputs are evaluated directly, without further search or assumed semantic compatibility.

We average over the $R\geq1$ retained references, $\bar v(S)=R^{-1}\sum_{r=1}^{R}v_r(S)$, and apply the Shapley--Taylor decomposition (Sec.~\ref{sec:background}). The signed main effects $\Phi_I$ and $\Phi_T$ measure score changes from restoring each modality alone; $\Phi_{IT}$ captures the residual joint effect. By linearity,
\begin{equation}
\Phi_I+\Phi_T+\Phi_{IT}
=\ell_{\hat y}(x)-\frac{1}{R}\sum_{r=1}^{R}\ell_{\hat y}(x'_r).
\label{eq:attr_efficiency}
\end{equation}
This identity holds exactly, including for a flagged fallback. Contributions depend on the search constraints, proposal and selection rules, and proximity criterion.

For visualization, we normalize attribution magnitudes:
\begin{equation}
s_M=\frac{100|\Phi_M|}{D},\qquad
D=\sum_{M\in\{I,T,IT\}}|\Phi_M|>0.
\label{eq:shares}
\end{equation}
These shares measure magnitude, not positive support for $\hat y$. Negative main effects indicate that restoring a modality alone lowers its log-odds. Negative and positive interactions indicate subadditivity and superadditivity, respectively. Absolute values hide these signs and any cancellation. When $D=0$, shares are undefined but the signed decomposition remains valid. We retain signed terms for distributional analysis and counterfactual edits for within-modality inspection.

\subsection{Attribution-Based Review}
\label{sec:method_tail}

For a fixed classifier and reference-generation configuration, define the signed attribution profile $\boldsymbol a(x)=(\Phi_I(x),\Phi_T(x),\Phi_{IT}(x))$, retaining interaction separately. A fitting population provides means $\mu_j$ and standard deviations $\sigma_j$ for standardizing nonconstant coordinates as $z_j(x)=(a_j(x)-\mu_j)/\sigma_j$. The same transformation applies to all inputs. Let $\mathcal A_{\mathrm{fit}}$ contain the standardized reference profiles. Define typicality as
\begin{equation}
\rho(x)=-d_k\!\left(\boldsymbol z(x),\mathcal A_{\mathrm{fit}}\right),
\label{eq:attribution_typicality}
\end{equation}
where $d_k$ is the Euclidean distance to the $k$-th nearest profile. Reusing the fitting inputs $X_1,\ldots,X_n$, compute leave-one-out scores
\[
\rho_j=-d_k\!\left(\boldsymbol z(X_j),\mathcal A_{\mathrm{fit}}^{(-j)}\right),
\qquad 1\leq k\leq n-1,
\]
where $\mathcal A_{\mathrm{fit}}^{(-j)}$ excludes entry $j$, while standardization remains fitted on all $n$ inputs. The empirical lower-tail rank is
\begin{equation}
\alpha(x)=\frac{1+\sum_{j=1}^{n}\mathbf 1\{\rho_j\leq\rho(x)\}}{n+1}.
\label{eq:tail}
\end{equation}
Inputs with $\alpha(x)\leq\alpha_0$ are prioritized for review alongside their edits and acceptance status. Reference scores use $n-1$ neighbors as candidates; evaluation scores use all $n$, so these ranks provide no split-conformal guarantee. Choosing $k$ above a shortcut cluster's size limits internal neighbor matching, but common or overlapping shortcuts may remain typical.

\section{Results}
\label{sec:results}

We evaluate CMA's modality attribution, counterfactual quality, shortcut detection and localization using controlled benchmarks.

\subsection{Experimental Setup}
\label{sec:setup}

\textbf{Controlled modality reliance.}
To enable quantitative evaluation of modality attribution, we construct three controlled multimodal benchmarks by pairing colored MNIST~\cite{lecun1998gradient} digits with textual prompts specifying a per-sample color-to-code mapping and name it Synthetic Multimodal MNIST. Synthetic Multimodal MNIST pairs a colored digit $a$ with a text record containing a number $b$, independently sampled with all $100$ classes equally represented. Three separate classifiers are trained to predict $a$ (Image-Biased), $b$ (Text-Biased), or the pair $(a,b)$ as one of $100$ classes (Balanced), requiring one or both modalities. Training, validation, and test sets contain $10{,}000/1{,}000/1{,}000$ multimodal inputs with disjoint MNIST images.

\textbf{Shortcut benchmarks.}
To evaluate atypicality detection and shortcut localization, we construct multimodal datasets with manually planted shortcuts and train classifiers to exploit them. For Synthetic MNIST, a classifier predicts ten digit classes: in the image-biased setting, a $3\times3$ corner marker indicates class $0$ in $2\%$ of inputs. In the text-biased setting, a class-specific \texttt{Site:} line indicates the label in $2\%$ of inputs, with other evidence randomized. For OpenI, we train a normal/abnormal classifier on $2{,}182$ studies. A burned-in \texttt{RVS} hospital marker serves as an image shortcut for the abnormal label in $6\%$ of studies.
\begin{table*}[htb]
\centering
\footnotesize
\setlength{\tabcolsep}{2.3pt}
\resizebox{\textwidth}{!}{%
\begin{tabular}{@{}l cc ccc cccc cccc@{}}
& \multicolumn{2}{c}{Detection}
& \multicolumn{3}{c}{Localization}
& \multicolumn{4}{c}{Image-biased classifier}
& \multicolumn{4}{c}{Text-biased classifier} \\
\cmidrule(lr){2-3}\cmidrule(lr){4-6}
\cmidrule(lr){7-10}\cmidrule(lr){11-14}
& \multicolumn{2}{c}{AUROC $\uparrow$}
& \multicolumn{2}{c}{MNIST}
& OpenI
& \multicolumn{4}{c}{Attribution and hit rate (\%)}
& \multicolumn{4}{c}{Attribution and hit rate (\%)} \\
Method
& MNIST & OpenI
& Image cue & Text cue & Image cue
& Img & Txt & Int & Hit $\uparrow$
& Img & Txt & Int & Hit $\uparrow$ \\
\midrule
Input$\times$Gradient
& 0.713 & 0.460 & 0.00 & 0.24 & 0.00
& 83.9\sd{0.8} & 16.1\sd{0.8} & -- & 94.3\sd{2.1}
& 77.3\sd{0.5} & 22.7\sd{0.5} & -- & 12.0\sd{1.0} \\
KernelSHAP
& 0.774 & 0.655 & 0.33 & 0.62 & 0.14
& 63.1\sd{0.1} & 36.9\sd{0.1} & -- & 99.7\sd{0.6}
& 51.6\sd{0.2} & 48.4\sd{0.2} & -- & 40.0\sd{0.0} \\
MM-SHAP
& 0.438 & 0.360 & 0.16 & 0.31 & 0.08
& 45.0\sd{0.7} & 55.0\sd{0.7} & -- & 28.0\sd{6.0}
& 35.4\sd{0.9} & 64.6\sd{0.9} & -- & 97.0\sd{2.0} \\
EAGLE
& 0.808 & 0.995 & 0.11 & \textbf{1.00} & 0.07
& 55.9\sd{0.3} & 44.1\sd{0.3} & -- & 67.7\sd{2.1}
& 30.7\sd{0.1} & 69.3\sd{0.1} & -- & 93.7\sd{0.6} \\
MultiViz
& 0.681 & 0.675 & 0.56 & 0.52 & 0.15
& 51.7\sd{0.4} & 48.3\sd{0.4} & -- & 57.3\sd{1.5}
& 26.5\sd{0.4} & 73.5\sd{0.4} & -- & \textbf{100.0}\sd{0.0} \\
InterSHAP
& 0.985 & \textbf{1.000} & -- & -- & --
& 91.2\sd{0.5} & 1.8\sd{0.2} & 7.0\sd{0.2}
& \textbf{100.0}\sd{0.0}
& 2.0\sd{0.5} & 91.9\sd{0.8} & 6.1\sd{0.4}
& \textbf{100.0}\sd{0.0} \\
SHAPE$^{\dagger}$
& -- & -- & -- & -- & --
& 10.0 & 2.2 & 87.8 & --
& 4.2 & 91.6 & 4.2 & -- \\
\midrule
\textbf{CMA}
& \textbf{0.987} & \textbf{1.000}
& \textbf{0.97} & \textbf{1.00} & \textbf{0.84}
& 87.7\sd{0.6} & 4.9\sd{0.6} & 7.3\sd{0.2}
& \textbf{100.0}\sd{0.0}
& 4.8\sd{0.2} & 88.5\sd{0.5} & 6.7\sd{0.5}
& \textbf{100.0}\sd{0.0} \\
\end{tabular}}
\caption{
\textbf{Modality attribution, shortcut detection, and localization.}
Detection reports atypicality AUROC for shortcut-following versus shortcut-free inputs, using shared reference populations and scoring across instance-level methods. Localization reports IoU with planted shortcuts. Attribution reports mean normalized magnitudes for image (Img), text (Txt), and interaction (Int) on controlled MNIST. Hit reports agreement (\%) between the top-ranked modality and ground truth modality. Attribution uses $100$ inputs per classifier. $\pm$ represents standard deviation over three explanation seeds. Dashes represent unavailable outputs. $^{\dagger}$SHAPE reports dataset-level scores.
}
\label{tab:attribution}
\end{table*}

\begin{table}[t]
\centering
\small
\setlength{\tabcolsep}{2pt}
\begin{tabular*}{\columnwidth}{@{\extracolsep{\fill}}lccccc@{}}
Dataset & Flip$\uparrow$ & SSIM$\uparrow$ & LPIPS$\downarrow$
        & BLEU-4$\uparrow$ & PPL$\downarrow$ \\
\midrule
Synthetic MNIST     & 100.0 & 0.867 & 0.078 & 0.852 & 69.8 \\
OpenI        &  98.5 & 0.951 & 0.036 & 0.678 & 88.3 \\
\end{tabular*}
\caption{
\textbf{Counterfactual quality.}
CMA changes predictions for nearly all evaluated inputs while retaining image structure and textual overlap.
Flip is the prediction-change rate (\%); SSIM and LPIPS (AlexNet) measure image similarity and perceptual distance, and BLEU-4 measures text overlap with the input.
PPL is GPT-2 perplexity of the edited text.
Results are averaged across three search seeds.
}
\label{tab:cf_quality}
\end{table}

\begin{table}[t]
\centering
\small
\setlength{\tabcolsep}{2pt}
\renewcommand{\arraystretch}{0.95}
\begin{tabular*}{\columnwidth}{@{\extracolsep{\fill}}clrrrrrr@{}}
& & \multicolumn{3}{c}{Attribution (\%)}
  & \multicolumn{3}{c}{Flagged (\%)} \\
\cmidrule(lr){3-5}\cmidrule(l){6-8}
& Shortcut & Img & Txt & Int & 5 & 10 & 20 \\
\midrule
\multirow{3}{*}{\rotatebox[origin=c]{90}{S. MNI.}}
& None  & 39.6 & 41.2 & 19.2 &  0.0 &   0.0 &   4.4 \\
& Image & 74.9 & 15.2 & 10.0 & 86.1 & 100.0 & 100.0 \\
& Text  &  8.0 & 72.9 & 19.1 & 61.9 &  73.8 &  81.0 \\
\midrule
\multirow{2}{*}{\rotatebox[origin=c]{90}{OpenI}}
& None  & 18.8 & 71.7 &  9.4 &  0.0 &   0.0 &   6.7 \\
& Image & 77.7 & 15.7 &  6.6 & 93.3 & 100.0 & 100.0 \\
\end{tabular*}
\caption{
\textbf{Attribution profiles and shortcut-induced attribution shifts.}
Img/Txt/Int are mean normalized attribution magnitudes.
Flagged is the percentage with empirical tail rank $\alpha(x)\leq\alpha_0$, where $\alpha_0$ is the review threshold.
At $\alpha_0=0.10$, CMA flags every evaluated image-shortcut case with no false alarms among the evaluated shortcut-free inputs.
Rows without shortcuts report false-positive rates.
Results average three explanation seeds after filtering shortcut cases for verified shortcut following.
}
\label{tab:tail}
\end{table}

\begin{table}[t]
\centering
\small
\begin{tabular*}{\columnwidth}{@{\extracolsep{\fill}}lrrr@{}}
& \multicolumn{3}{c}{Mean CMA attribution (\%)} \\
\cmidrule(l){2-4}
Case-text length & Image & Text & Interaction \\
\midrule
Short ($n=16$)    & 41.8 & 29.8 & 28.4 \\
Medium ($n=16$)  & 24.8 & 36.9 & 38.3 \\
Long ($n=15$)   & 16.1 & 52.7 & 31.2 \\
\end{tabular*}
\caption{
\textbf{Modality attribution through a commercial MLLM API.}
Mean CMA attribution magnitudes for $47$ NEJM dermatology cases evaluated with Qwen3.8 Omni Flash via OpenRouter.
As more text is provided with information CMA correctly quantifies modality attribution shift from image to text.
}
\label{tab:api_nejm}
\end{table}

\textbf{Classifiers and behavioral checks.}
Each controlled classifier combines a pretrained Gemma-3-4B-IT backbone~\citep{gemma3team2025} with a two-layer MLP trained on its final-token representation. Test accuracies are $95.4\%$, $100\%$, and $95.0\%$ for Image-Biased, Text-Biased, and Balanced MNIST, respectively, and $97.7\%$ and $98.6\%$ for the MNIST and OpenI shortcut benchmarks
Shortcut evaluation retains inputs whose predictions follow a planted shortcut despite contradictory evidence. Shortcut-following rates are $83\%$ for the MNIST marker, $77\%$ for the site line, and $97\%$ for the OpenI hospital.

\textbf{Comparison methods.}
We compare with Input$\times$Gradient~\citep{ancona2017towards}, which weights input gradients by input values; KernelSHAP~\citep{lundberg2017shap}, which estimates Shapley attributions from perturbations; and MM-SHAP~\citep{parcalabescu2023mm}, which aggregates token and image-region contributions into modality scores. EAGLE~\citep{chen2026mllms} searches for sufficient and necessary visual evidence, while MultiViz~\citep{liang2022multiviz} analyzes unimodal importance and cross-modal interactions. We aggregate input-level scores by modality where required. InterSHAP~\citep{wenderoth2025measuring} separates modality contributions from interactions, and SHAPE~\citep{hu2022shape} measures dataset-level contribution and cooperation through predictive performance.

\textbf{Evaluation metrics.}
Attribution shares normalize absolute contributions (Eq.~\ref{eq:shares}). The interaction share captures non-additive image--text effects. Hit rate measures agreement between the top-ranked and ground-truth modalities. Detection AUROC measures how well atypicality separates shortcut-following from shortcut-free inputs. Localization IoU measures overlap between the planted shortcut and the attributed or CMA-edited region. Counterfactual flip rate is the proportion of inputs whose predictions change.
Structural similarity (SSIM) and learned perceptual image patch similarity (LPIPS) assess image proximity. BLEU-4 measures text overlap, and GPT-2 perplexity (PPL) measures edited-text likelihood under a reference language model. Higher hit rate, AUROC, IoU, flip rate, SSIM, and BLEU-4 are preferred. Lower LPIPS and PPL are preferred.

\subsection{Modality Attribution with Measured Reliance}
\label{sec:exp_attr}

We test whether explanations recover verified image or text dependence on $100$ inputs per biased classifier.
We evaluate $100$ inputs per biased classifier, defining a hit as agreement with the ground truth modality.
CMA identifies the dominant modality for every input evaluated in three seeds, assigning $87.7\%$ to images in the Image-Biased setting and $88.5\%$ to text in the Text-Biased setting (Tab.~\ref{tab:attribution}). The reversal of dominant attribution across tasks shows that CMA follows the classifier's reliance rather than favoring one modality.
Among the input-level methods, InterSHAP also achieves $100\%$ but provides weaker localized evidence.

\subsection{Counterfactual Quality}
\label{sec:exp_cfq}

We assess the quality of counterfactuals generated by CMA (Tab.~\ref{tab:cf_quality}).
Flip rates reach $100\%$ on MNIST and $98.5\%$ on OpenI, with SSIM of $0.867/0.951$ and LPIPS of $0.078/0.036$, showing that prediction changes are achieved without large changes. MNIST perplexity remains close to the original with $69.8$ versus $69.5$, OpenI edits yield BLEU-4 of $0.678$ and GPT-2 perplexity of $88.3$, indicating little change in language-model likelihood. These results support CMA's ability to produce effective counterfactuals while preserving measured visual and textual content.

\subsection{Detecting Shortcut-Driven Predictions}
\label{sec:exp_tail}

\textbf{Detection protocol.}
We test whether shortcut-driven predictions depart from the model's usual modality reliance distribution. We standardize each signed attribution profile against a reference population, retaining image, text, and interaction as separate coordinates (Sec.~\ref{sec:method_tail}). A nearest-neighbor score measures atypicality using $k=25$ reference neighbors.%
Reference populations contain $200$ MNIST training inputs and $100$ OpenI inputs, including four examples per shortcut type. Leave-one-out scores within these populations calibrate the empirical tail rank $\alpha(x)$ (Eq.~\ref{eq:tail}): smaller ranks indicate more unusual reliance. We flag input $x$ when $\alpha(x)\leq\alpha_0$, where $\alpha_0$ is the review threshold. All methods use the same reference inputs and scoring procedure. Candidate evaluation pools contain $15$ inputs per MNIST group and $10$ per OpenI group before filtering shortcut cases for verified shortcut following.

\textbf{Shortcut detection.}
CMA exposes systematic shifts toward the shortcut-bearing modality (Tab.~\ref{tab:tail}). The MNIST image marker raises mean image attribution from $39.6\%$ to $74.9\%$, while the \texttt{Site:} line raises text attribution from $41.2\%$ to $72.9\%$. In OpenI, the planted hospital token reverses the dominant modality from text ($71.7\%$ without the shortcut) to image ($77.7\%$ with it). These shifts make shortcut reliance detectable from the attribution profile.
At a review threshold of $\alpha_0=0.10$, CMA flags every evaluated image-shortcut case and $73.8\%$ of text-shortcut cases, with no false alarms among the evaluated shortcut-free inputs. Increasing the threshold to $0.20$ increases the detection of text shortcuts to $81.0\%$, with false-positive rates of $4.4\%$ on MNIST and $6.7\%$ on OpenI. CMA achieves AUROC detection of $0.987/1.000$ on MNIST/OpenI, compared with $0.985/1.000$ for InterSHAP and $0.808/0.995$ for EAGLE (Tab.~\ref{tab:attribution}).  CMA pairs highly accurate shortcut detection with counterfactual localization to inspect flagged predictions.%

\begin{figure*}[t!]
  \centering
  \begin{subfigure}[b]{0.24\textwidth}
    \includegraphics[width=\linewidth]{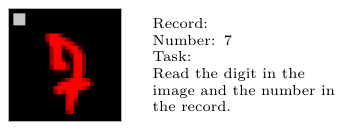}
    \caption{Original (pred = 0)}
  \end{subfigure}\hfill
  \begin{subfigure}[b]{0.24\textwidth}
    \includegraphics[width=\linewidth]{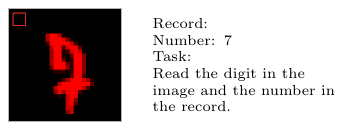}
    \caption{Image-only CF (pred = 7)}
  \end{subfigure}\hfill
  \begin{subfigure}[b]{0.24\textwidth}
    \includegraphics[width=\linewidth]{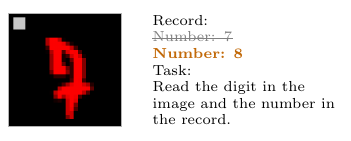}
    \caption{Text-only CF (pred = 0)}
  \end{subfigure}\hfill
  \begin{subfigure}[b]{0.24\textwidth}
    \includegraphics[width=\linewidth]{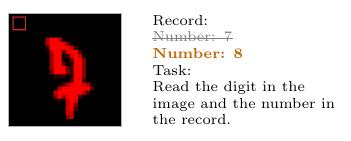}
    \caption{Joint CF (pred = 8)}
  \end{subfigure}
  \vskip 0.5em
  \begin{subfigure}[b]{0.24\textwidth}
    \includegraphics[width=\linewidth]{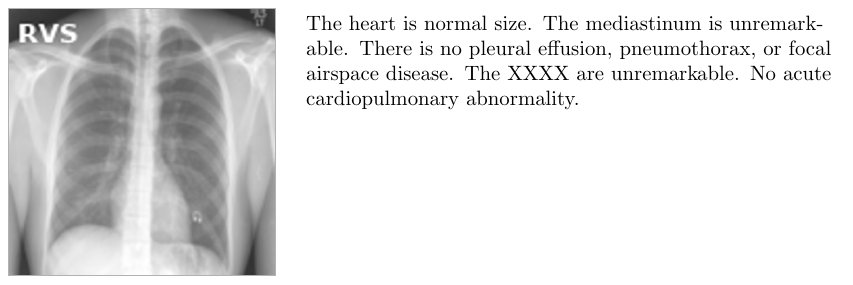}
    \caption{Original (Abnormal)}
  \end{subfigure}\hfill
  \begin{subfigure}[b]{0.24\textwidth}
    \includegraphics[width=\linewidth]{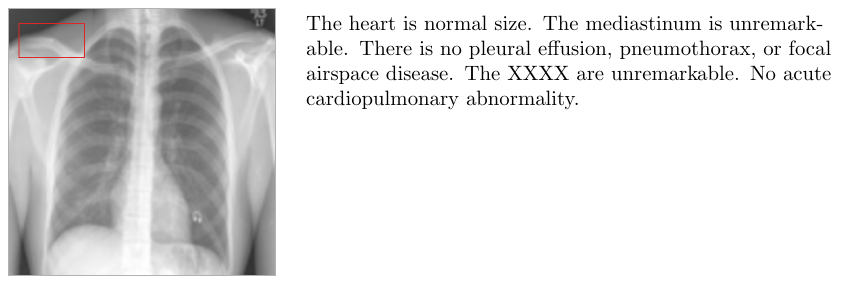}
    \caption{Image-only CF (Normal)}
  \end{subfigure}\hfill
  \begin{subfigure}[b]{0.24\textwidth}
    \includegraphics[width=\linewidth]{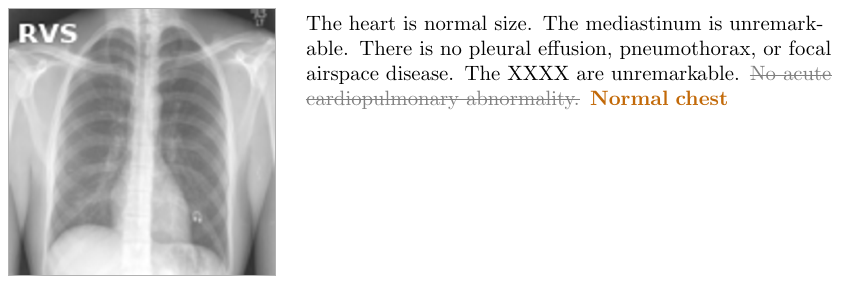}
    \caption{Text-only CF (Abnormal)}
  \end{subfigure}\hfill
  \begin{subfigure}[b]{0.24\textwidth}
    \includegraphics[width=\linewidth]{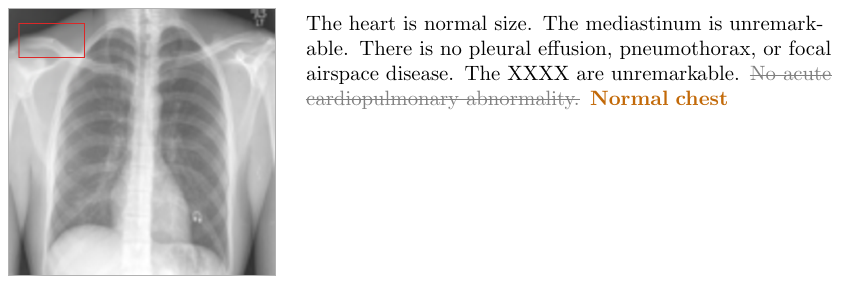}
    \caption{Joint CF (Normal)}
  \end{subfigure}
  \caption{
  \textbf{Counterfactual inspection of image shortcuts.}
 Top: a planted corner marker causes a digit $7$ to be classified as $0$. Removing the marker restores prediction $7$; the text-only edit leaves it at $0$, and the joint edit yields $8$. Bottom: removing the planted \texttt{RVS} hospital token changes an OpenI prediction from abnormal to normal; the text-only edit leaves it abnormal. Red boxes mark image edits; amber and grey mark inserted and deleted text. These examples link atypical image reliance to localized prediction-changing edits.
}
  \label{fig:cf_modes}
\end{figure*}

\textbf{Shortcut localization.}
We test whether explanations identify the planted content responsible for shortcut reliance. CMA achieves image-shortcut IoU of $0.97$ on MNIST and $0.84$ on OpenI, exceeding the strongest compared input-level method by $0.41$ and $0.69$, respectively (Tab.~\ref{tab:attribution}). For the textual \texttt{Site:} shortcut, CMA matches EAGLE's  IoU of $1.00$. CMA therefore localizes both visual and textual shortcuts.
Fig.~\ref{fig:cf_modes} illustrates counterfactual inspection of image shortcuts. Removing the MNIST marker changes the prediction from $0$ to the depicted digit $7$; removing the OpenI token changes abnormal to normal. The displayed text-only edits leave both predictions unchanged. These interventions link atypical image reliance to specific prediction-changing edits, making shortcuts inspectable.

\subsection{Attribution through a Commercial Model API}
\label{sec:exp_api}

We evaluate CMA on a native autoregressive MLLM accessed through a commercial API, using Qwen3.8 Omni Flash via OpenRouter on NEJM dermatology cases~\citep{buckley2026multimodal}. Each case contains an image, a description, and five answer options. CMA explains the selected option's log-odds against the other four using API-returned first-answer-token probabilities. We use $47$ cases that are divided into approximately equal groups by description word count.
Across the shortest to longest groups, mean text attribution rises from $29.8\%$ to $52.7\%$, while image attribution falls from $41.8\%$ to $16.1\%$ (Tab.~\ref{tab:api_nejm}). Text attribution is positively associated with description length (Spearman rank correlation $\rho=0.36$, $p$-value $0.012$), indicating greater textual reliance among cases with longer descriptions. Interaction accounts for $28.4$--$38.3\%$ of attribution magnitude, exposing a substantial non-additive contribution alongside the modality-specific effects.
This association aligns with the text reliance reported by \citet{buckley2026multimodal} showing that as we provide more information in the text LLM uses it over image which is correctly quantified by the CMA.

\subsection{Progressive Modality Corruption}
\label{sec:exp_ablation}

\begin{figure*}[t]
\centering
\includegraphics[width=0.95\linewidth]{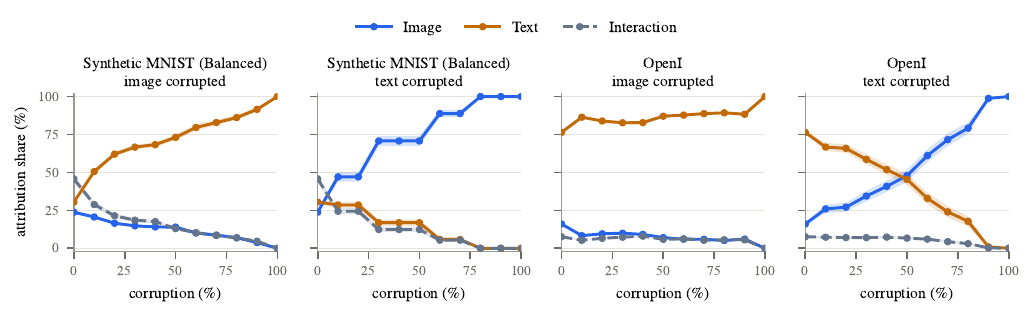}
\caption{
\textbf{Modality attribution under progressive modality corruption.}
Gemma-3-4B results on Balanced MNIST and shortcut-free OpenI, averaged over three seeds per input. As corruption increases, attribution shifts toward the intact modality and the interaction share declines, demonstrating sensitivity to input corruption rather than establishing faithfulness.
}
\label{fig:corruption}
\end{figure*}

We perform a controlled per-modality corruption study to evaluate whether CMA tracks the evidence available to the model rather than assigning a fixed preference to either modality. Image and text are progressively corrupted from $0\%$ to $100\%$ in $10\%$ increments. Image corruption progressively replaces pixels with the background value, while text corruption masks increasing numbers of tokens through the attention mechanism.
Fig.~\ref{fig:corruption} presents the results on OpenI and the balanced Synthetic MNIST benchmark. As information in one modality is progressively removed, its attribution decreases toward $0\%$, while the complementary modality receives the remaining attribution. CMA’s attribution changes as modality content is progressively corrupted. When full corruption makes one modality identical across coalitions, its main effect and interaction vanish by construction, leaving the intact modality with $100\%$ of the attribution magnitude when the remaining contribution is nonzero. This reflects the remaining score contrast, not necessarily stronger evidence or a correct prediction. Modality attribution thus complements accuracy by showing how the distribution of contributions changes under corruption.

\section{Discussion}
\label{sec:discussion}

CMA links modality attribution to localized prediction-changing edits. Identifying the dominant modality alone cannot distinguish relevant evidence from shortcuts. Attribution profiles prioritize decisions for review, while counterfactual expose influential content. Experiments demonstrate accurate dominant-modality identification, shortcut detection, and localization, supporting the inspection.

\textbf{Interpreting modality contributions.}
Separating image, text, and interaction terms distinguishes modality-specific from non-additive effects. Under corruption, attribution shifts to the intact modality as interaction declines. These trends describe changes in counterfactual score contrasts. They do not establish preserved accuracy or faithful reasoning. Interpretation depends on the  references.

\textbf{Using attribution for shortcut review.}
Atypical profiles can flag shifts in modality reliance, but legitimate inputs may also appear unusual, while common shortcuts may remain undetected. Counterfactual edits support inspection without certifying correctness.

\textbf{Limitations.}
CMA depends on diffusion priors, editing constraints, and search success. Prediction flips, similarity, and reconstruction stability do not guarantee semantic or clinical validity. Attributions describe model-relative contrasts, not data-generating causal effects. Evidence comes primarily from controlled classifiers with planted shortcuts. Review thresholds remain empirical because fitting and calibration reuse the same population. Generation takes $45$--$71$\,s per input on one GPU, favoring offline auditing.

\section{Conclusion}
\label{sec:conclusion}
We introduced Counterfactual Modality Attribution (CMA) for investigating shortcut-driven predictions. Experiments show accurate modality identification and shortcut detection, with improved localization over evaluated baselines. Beyond inspection, reliance profiles could support active learning, dataset auditing, and deployment monitoring. Future work should test whether atypical reliance identifies confident errors missed by uncertainty-based selection or detects behavioral shifts before performance declines.

\begin{figure}[t]
\section*{LLM Usage Statement}
AI-assisted tools were used to assist with manuscript editing, literature organization, figure refinement, and data preparation, as well as to support code development and experimental workflows. The authors reviewed and verified all AI-assisted outputs and retain full responsibility for the research, its findings, and the submitted content.
\end{figure}

{
    \small
    \bibliographystyle{ieeenat_fullname}
    \bibliography{main}
}

\end{document}